\documentclass[10 pt,journal,twoside]{IEEEtran}

\usepackage{silence}
\usepackage{mathptmx}
\usepackage{amsmath}
\usepackage{amssymb}
\usepackage{algpseudocode}
\usepackage{array}
\usepackage{graphicx}
\usepackage{tabularx}
\graphicspath{ {Images/} }
\usepackage{multirow}
\usepackage{setspace}
\usepackage[style=base]{caption}
\usepackage{subcaption}
\usepackage{url}
\usepackage{xcolor}
\usepackage{multirow}

\usepackage{times}
\usepackage{fancyhdr,graphicx,amsmath,amssymb}
\usepackage[ruled,vlined,linesnumbered]{algorithm2e}
\usepackage[utf8]{inputenc}
\usepackage[backend=biber,style=ieee,sorting=none]{biblatex}
\usepackage{eucal}

\begin{document}
\title{An Intelligent Passive Food Intake Assessment System with Egocentric Cameras}

\author{Frank Po Wen Lo,~\IEEEmembership{Student Member,~IEEE,}
         Modou L Jobarteh,
         Yingnan Sun,~\IEEEmembership{Member,~IEEE,}
         Jianing Qiu,~\IEEEmembership{Student Member,~IEEE,}
         Shuo Jiang,~\IEEEmembership{Student Member,~IEEE,}
         Gary Frost,
         and~Benny Lo,~\IEEEmembership{Senior Member,~IEEE}% <-this % stops a space 

\thanks{This project is supported by the Innovative Passive Dietary Monitoring Project funded by the Bill \& Melinda Gates Foundation (OPP1171395).}
\thanks{Frank Po Wen Lo, Yingnan Sun and Benny Lo are with the Department
of Surgery and Cancer, the Hamlyn Centre, Imperial College London, London,
United Kingdom, e-mail: \{po.lo15, y.sun16, benny.lo\}@imperial.ac.uk }% <-this % stops a space
\thanks{Jianing Qiu is with the Department of Computing, the Hamlyn Centre, Imperial College London, London,
United Kingdom, e-mail: \ jq916@ic.ac.uk }
\thanks{Modou L Jobarteh and
        Gary Frost are with the Department of Metabolism, Digestion and Reproduction, Imperial College London, London,
United Kingdom, e-mail: \{m.jobarteh, g.frost\}@imperial.ac.uk }
\thanks{Shuo Jiang is with the State Key Laboratory of Mechanical System and Vibration, School of Mechanical Engineering, Shanghai Jiao Tong University, Shanghai, China, e-mail: jiangshuo@sjtu.edu.cn}
}


\markboth{Journal of \LaTeX\ Class Files,~Vol.~14, No.~8, August~2015}%
{Shell \MakeLowercase{\textit{et al.}}: Bare Demo of IEEEtran.cls for IEEE Journals}




\maketitle


\begin{abstract}

Malnutrition is a major public health concern in low-and-middle-income countries (LMICs). Understanding food and nutrient intake across communities, households and individuals is critical to the development of health policies and interventions. To ease the procedure in conducting large-scale dietary assessments, we propose to implement an intelligent passive food intake assessment system via egocentric cameras particular for households in Ghana and Uganda. Algorithms are first designed to remove redundant images for minimising the storage memory. At run time, deep learning-based semantic segmentation is applied to recognise multi-food types and newly-designed handcrafted features are extracted for further consumed food weight monitoring. Comprehensive experiments are conducted to validate our methods on an in-the-wild dataset captured under the settings which simulate the unique LMIC conditions with participants of Ghanaian and Kenyan origin eating common Ghanaian/Kenyan dishes. To demonstrate the efficacy, experienced dietitians are involved in this research to perform the visual portion size estimation, and their predictions are compared to our proposed method. The promising results have shown that our method is able to reliably monitor food intake and give feedback on users' eating behaviour which provides guidance for dietitians in regular dietary assessment.

\end{abstract}


\section{Introduction}

As part of its 2016-2025 nutrition strategy, the World Health Organisation (WHO) collaborates with member states to ensure universal access to healthy diets and effective interventions for ending all forms of malnutrition. This target is of particular importance to sub-Saharan Africa where populations are undergoing a nutrition transition \cite{whowebsite}. Despite a considerable portion of populations suffering from undernutrition in Africa, overweight is increasing in certain urban areas along with the burden of diet-related diseases, contributing to the so-called double burden of malnutrition \cite{jobarteh2020development}. Since food intake is an important contributor to the malnutrition burden, understanding the type, quantity and nutritional content of habitual diets eaten at the community, household and individual level is critical to the formulation of policies and interventions to eradicate or reduce malnutrition.
\begin{figure}
\centering
\includegraphics[width=\columnwidth]{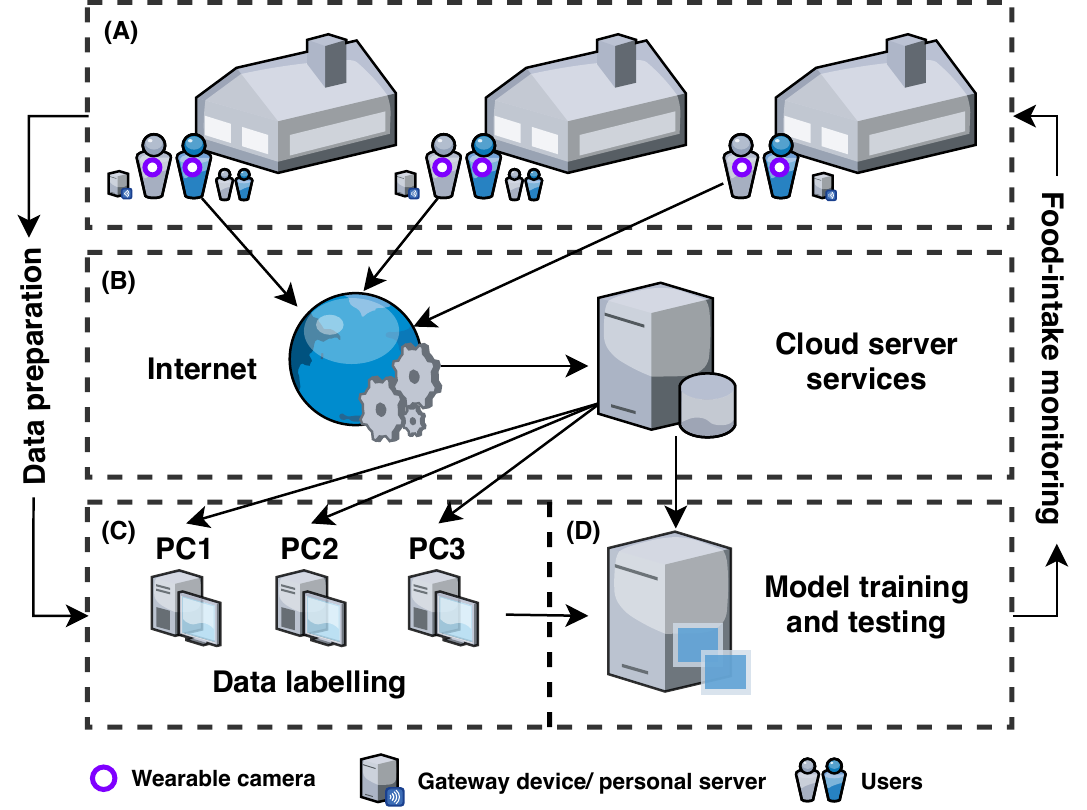}
\caption{An illustration diagram of the passive food intake assessment system for large-scale dietary assessments. (A) Household dietary data
collection; (B) Data transmission and cloud storage; (C) Data
labelling; (D) Model training and testing. Note that the cameras/gateway devices will be distributed, set up and collected by our staff in the field site}
\label{flowchart}
\end{figure}
\begin{figure*}[htb]
\centering
\includegraphics[width=\textwidth]{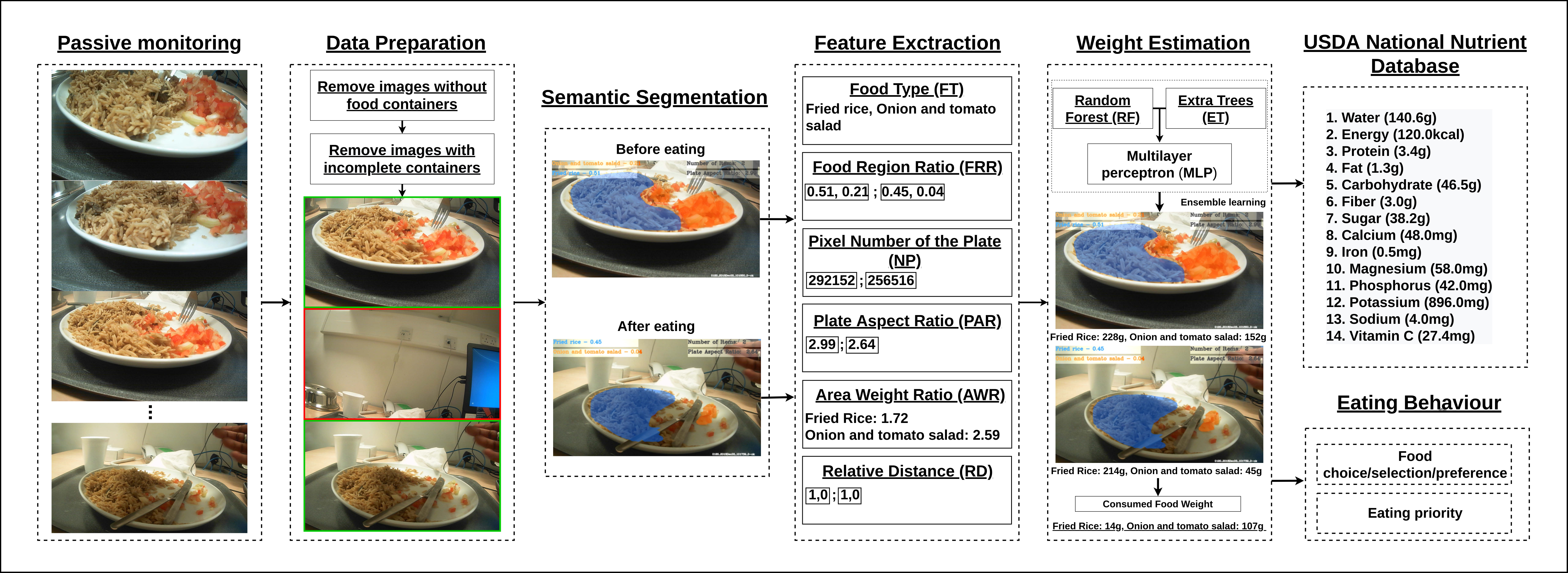}
\caption{A detailed implementation of food segmentation and consumed food weight estimation}
\label{detailedpipeline}
\vspace{-15pt}
\end{figure*}
A dietary assessment method, known as 24-h recall (24HR), has been the mainstay of nutritional epidemiology studies. However, this self-reporting method relies on an individual’s memory to estimate the quantities of foods and drinks consumed in a 24-hour period, contributing to reporting bias and erroneous conclusions. This subjective method is also labour intensive, expensive, and often difficult to conduct large-scale assessments. Efforts have been made in high-income countries by using various vision-based dietary assessment techniques to minimise reporting bias and improve the efficiency of data capture, e.g., actively capture an image of the meal by a smartphone and send it to the server for automatic analysis \cite{zhu2014multiple,zhu2010image,zhu2010use,meyers2015}. However, such methods require access to computer literacy and numeracy skills, which limit the deployability in rural regions. Without relying on the active method which requires certain human intervention, e.g., take photos and self-report, passive wearable technology \cite{selamat2020automatic}, such as the piezoelectric sensor \cite{alshurafa2015recognition,farooq2016segmentation}, hand gesture sensor \cite{salley2016comparison}, acoustic sensor \cite{mirtchouk2016automated,turan2018detection,bi2015autodietary}, accelerometer \cite{farooq2018accelerometer,arun2020accelerometer} and electromyography (EMG) sensor \cite{huang2017your,zhang2017monitoring}, provides user-friendly alternatives to conduct large-scale dietary intake assessments in LMIC regions. Although useful information about food intake, e.g., bite count rates and meal duration \cite{shen2016assessing}, can be obtained, there are still limitations on these existing passive methods. For instance, the food classification capability is limited that only the broad food categories can be recognised \cite{huang2017your}. Most importantly, it is challenging to accurately quantify the consumed portion size by just counting the bites \cite{amft2009bite,qiu2020counting}. 

To address these problems, our group proposed a passive food intake assessment system for wearable cameras to help reduce the burden and technical requirements for users in dietary information collection and analysis in LMIC regions especially for annual nutrition survey. Our vision-based food intake assessment system is developed by incorporating data-driven semantic segmentation and machine learning approaches. Not only does semantic segmentation recognises the food contents, but its region proposal property also locates the food objects. Then feature engineering is exploited to extract several new handcrafted features particularly designed for food weight estimation task. To evaluate the performance of our proposed method, comprehensive experiments have been done on a large-scale in-the-wild dataset with foods of Ghanaian and Kenyan origin captured under LMIC specific conditions, such as dinning in a well-lit or low-lit room to mimic a scenario of inadequate electricity supply and using shared plates to mimic a communal eating scenario. Unlike previous active capture approaches (e.g., take a snap shot), passive monitoring will generate a vast number of images, out of which more than 90\% of images contain no food or eating instance. Because of the passive nature of the camera devices, it is also not uncommon for them to capture distorted food images (i.e., only parts of the food containers are visible in the images); removing these artefacts prior to analysis improves the accuracy of the food weight estimation. Thus, in addition to the weight estimation algorithm, we also developed new data preparation techniques to improve analysis of food images captured in passive food intake monitoring. To further demonstrate the efficacy of our system, experienced dietitians are involved in this research to perform the visual food weight estimation, and their predictions are compared to our proposed method. The main contributions of this paper can be summarised as follows: (1) An intelligent passive assessment system is developed to continuously monitor the dietary-intake (food content and weight) and give feedback; (2) The weight of food remains can be estimated using our method such that the consumed food weight can be computed instead of just estimating initial food weight proposed in previous works; (3) Novel handcrafted features are designed and extracted particularly for food weight estimation. Various ML methods (e.g., Random forest, gradient boosting, decision tree and neural network) are applied to evaluate the efficiency of those features. Feature importance is calculated to show how each individual feature contributes to the final estimation; (4) Data preparation algorithms are designed for wearable cameras to provide a potential solution to remove redundant frames and images with incomplete food containers; (5) A food dataset with actual weight labelled is constructed to train and evaluate the proposed food weight estimation approach in the wild.

%(4) Data preparation algorithms are investigated for wearable cameras to remove redundant frames and images with incomplete food containers.

\section{Related Works}

To accurately quantify an individual's food intake, measuring the actual amount of foods consumed (portion size) is critical. With the recent development in computer vision techniques, different vision-based methods have been proposed to solve the problem of portion size estimation \cite{lo2019point2volume}. Specifically, the methods can be divided into several main categories ranging from stereo-based \cite{dehais2017,rahman2012food}, model-based \cite{7442364,jia2014accuracy,xu2013} and perspective transformation \cite{yang2019image,jia2012,lo2020image} approach. Although these methods show promising results in portion size estimation, they are mostly designed for methods which use active capture of food intake (e.g., mobile phone-based methods), and thus difficult to implement in studies using passive capture of food intake with wearable cameras. For instance, stereo-based method has several constraints on the capturing angles, e.g., at least two images should be taken from different viewing angles \cite{lo2020image}, to achieve feature matching and 3D reconstruction which in turn makes this method difficult to be used along with wearable cameras mounted on certain specific locations. Another concern is that stereo-based approach relies heavily on feature matching between frames in order to obtain the 3D geometry of the food items. However, under conditions of low lighting in LMIC, the food surface will not show distinctive texture and characteristics which makes the situation even harder. For model-based approach, it calculates the food volume by matching the food items in the image with the pre-build 3D food models. For instance, \cite{sun2015} proposed a virtual reality approach by superimposing 3D models with known volume onto the food items. However, this method requires certain human intervention such as rotating and scaling the 3D models which makes the method only suitable for use in active capture of food intake. Most importantly, the major concern is that model-based approach only measures initial portion size (i.e., the quantity of foods at the start of eating). It does not provide an estimation of the consumed portion size (i.e., the total quantity of foods eaten) contributing to large errors in portion size estimation. Similar to stereo-based approach, perspective transformation approach also requires users to use hand-held cameras and take images from specific locations, e.g., putting the mobile phones on the table \cite{yang2019image}, capturing images using bird-eye view \cite{lo2019depth}. Compared to previous works, our proposed method can provide food weight estimation without requiring specific image capturing angles and positions. To the best of our knowledge, there are limited works focusing on passive vision-based monitoring systems that estimate the consumed food weight continuously. Instead, most of the previous studies measure the portion size of the whole meal by just taking a snapshot. Due to the continuous capture of our system, researchers are able to investigate into the eating behaviour of the households in Ghana and Kenya (e.g., prefer eating starchy foods first) and facilitate the implementation of effective nutrition actions and policies in close collaboration with local governments.

\section{Detailed Information and Methods}
\subsection{System Modelling}

Figure \ref{flowchart} shows a three-tier Body Sensor Network (BSN) based passive food intake assessment system, in which the images or videos captured by wearable cameras are forwarded to cloud servers for data storage. In urban regions, NVIDIA Jetson nano, an embedded platform, can be used as a suitable gateway device due to its physical size, price and AI compute density \cite{dorrer2020comparison,de2020vineyard}. With the support of NVIDIA CUDA and TensorRT software libraries, deep learning frameworks can be implemented on such embedded platforms to make it easy to carry out AI applications and data preparation on site. Specifically, our proposed data preparation algorithms can significantly reduce the data to be transmitted to the cloud servers. Note that the wearable cameras/gateway devices will be distributed, set up and collected by our staff in the field site. However, the urban-rural divide in LMIC complicates the implementation of the system. In consideration of the low internet availability and inadequate electricity supply in rural regions, an alternative approach is proposed which is to temporarily store captured images to the on-board SD card, and synchronise with the gateway device when available or manually collect by our staff and upload to the cloud server after going back to urban regions. After uploading the images, experienced dietitians, located in London, will access the cloud server for data labelling, which includes food recognition and portion size estimation, using the image annotation software, such as Automatic Ingestion Monitor (AIM) software \cite{jobarteh2020development} and VGG Image Annotator \cite{dutta2019vgg}. Then the annotated images are forwarded to train a semantic segmentation and a portion size estimation network for food segmentation and consumed food weight estimation respectively (training process). Once the models are trained, the uploaded images will be forwarded to the trained networks for automatic dietary assessments (testing process), as shown in Figure \ref{detailedpipeline}.

\begin{figure}[t]
\centering
\includegraphics[width=\columnwidth]{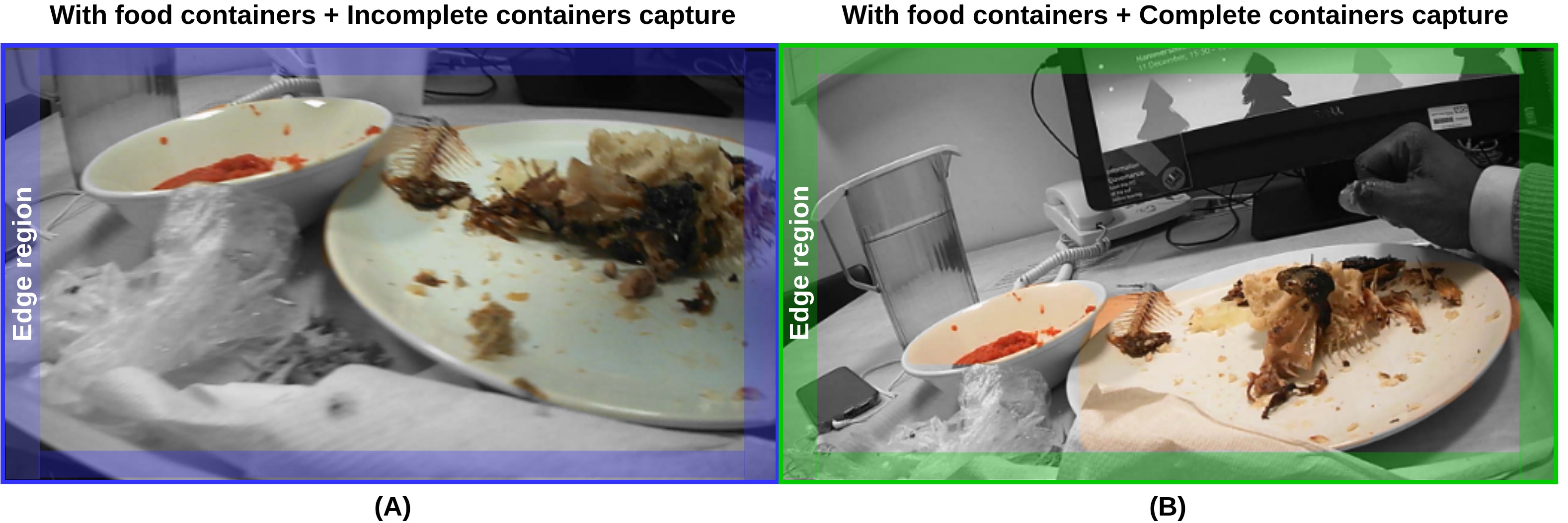}
\caption{An illustration diagram of the proposed method to remove images with incomplete food containers. (A) High proportion of overlapping region (B) Low proportion of overlapping region between the segmentation masks of the food containers and the edge region}
\label{edge}
\vspace{-10pt}
\end{figure}

\begin{figure*}[t]
\centering
\includegraphics[width=\textwidth]{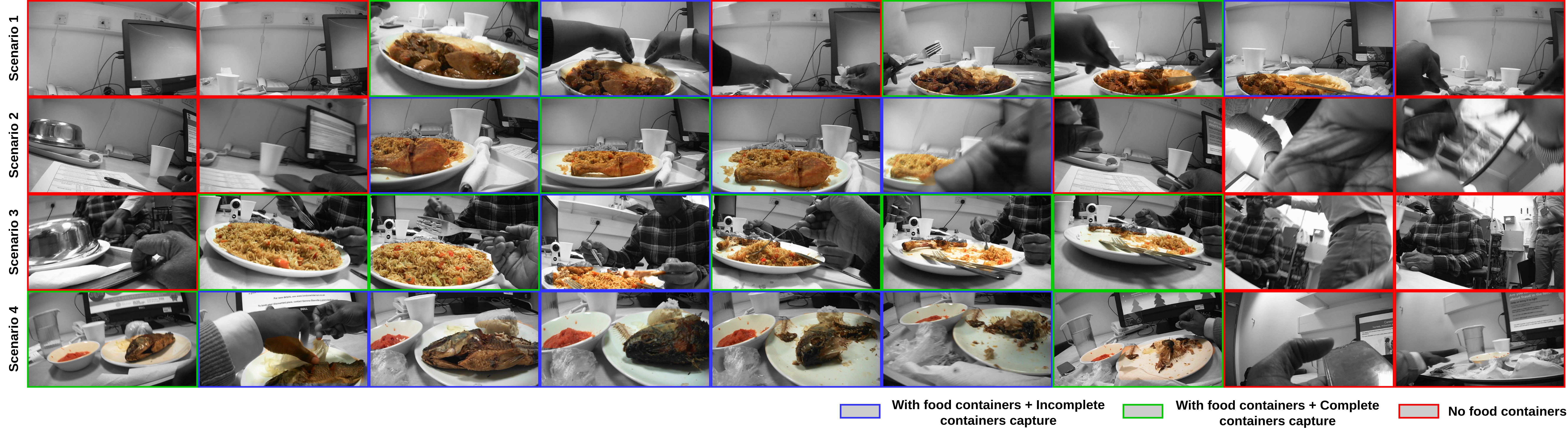}
\caption{Sample images obtained using the proposed data preparation methods in different scenarios; with food containers but incomplete containers capture (blue), with food containers and complete containers capture (green), no food containers (red)}
\label{datapreparation}
\vspace{-10pt}
\end{figure*}

\subsection{Data preparation}
Despite the convenience of the passive method, the implementing strategies are relatively complicated since the data storage for continuous long-term surveillance is a challenging technical issue. It is because the passive method will generate image frames continuously, which requires a huge memory space of the devices. Several previous studies proposed selecting representative frames by circular plate detection using Canny detector \cite{sun2015,nie2010}. Recently, \cite{jia2019automatic} proposed using Clarifai CNN, which output tags for each input image and determine the importance by counting the food-related tags. Similar to \cite{jia2019automatic}, we also proposed a deep learning approach to determine whether food items are presented in the images or not for removing redundant frames. The major difference is that we applied semantic segmentation to segment and locate the food containers instead of just using a classifier to recognise which frame is important. When the image pixel number of food containers is larger than a pre-set threshold, this image will be considered as a representative frame. The details of the proposed method can be found in Algorithm \ref{algorithm1} and the architecture of the segmentation network will be shown in the following section.

\setlength{\intextsep}{0pt} 
\begin{algorithm}

\SetAlgoLined
\KwData{\textit{frame[i]} refers to each consecutive frame captured by wearable cameras; \textit{NP[i]} refers to the number of pixels of the food container in \textit{frame[i]} obtained through the semantic segmentation; \textit{threshold} represents the minimum number of pixels to determine the occurrence of food containers}
\KwResult{\textit{frames\_container} is a list which stores the index of the frames with food containers}
 \For{\textit{i}=0; \textit{i}$<$\textit{frame\_number}; \textit{i++}}{
 \If{\textit{NP[i]}$>$\textit{threshold}}
 {append \textit{i} to \textit{frames\_container}}
 }
 \caption{Redundant images removal algorithm}
 
 \label{algorithm1}
\end{algorithm}

In addition to removing images with no food containers, we also proposed a method to further remove images where the full food container is not visible (i.e., images with incomplete food container). These removals are critical as our proposed food weight estimation network relies on the proportion between the food container and the food items on it. Conducting the estimation on images where only parts of the food container are visible increases the proportion of food items which in turn causes overestimation of the food weight. In developing the method, we first generate the segmentation masks of the food containers for each image obtained from the previous step. Then an alterable edge region is defined, as shown in Figure \ref{edge} (the region in blue and green colour), of which the parameter (e.g., the width of the edge region) could be changed according to the resolution of the images captured by wearable cameras (e.g., $640\times480$, $1280\times720$). When a high proportion of the area by the segmentation mask of the food container and edge region is in overlap, the corresponding image will be considered as containing an incomplete food container. The detailed implementation of the method is presented in Algorithm \ref{algorithm2} and more experimental results of our proposed methods are shown in Figure \ref{datapreparation}.

\begin{algorithm}
\SetAlgoLined
\KwData{\textit{mask[i]} refers to segmentation mask of the food container in \textit{frame[i]}. \textit{edge\_counter[i]} stores the number of pixels that the mask of the container overlaps with the edge region. \textit{threshold} represents the minimum number of pixels to determine the incompleteness of food containers}
\KwResult{\textit{frames\_container} is a list which stores the index of the frames with complete food containers}
 %initialization\;
 \For{\textit{i}=0; \textit{i}$<$\textit{mask\_number}; \textit{i++}}{
 \For{each pixel in \textit{mask[i]}}
 {\If{the pixel overlaps with the edge region}
    {\textit{edge\_counter[i]++}}}
 
%  \newline 
% \textbf{loop} to determine how many pixels in \textit{mask[i]} overlap with the edge region
 
\If{\textit{edge\_counter[i]++}$<$\textit{threshold}}
 {append \textit{i} to \textit{frames\_container}}}
 \caption{Remove images containing incomplete food containers}
\label{algorithm2}
\end{algorithm}
\setlength{\textfloatsep}{5pt}
%%traverse every point

\vspace{-5pt}
\subsection{Semantic segmentation network}

\begin{figure}[tb]
\centering
\includegraphics[width=0.95\columnwidth]{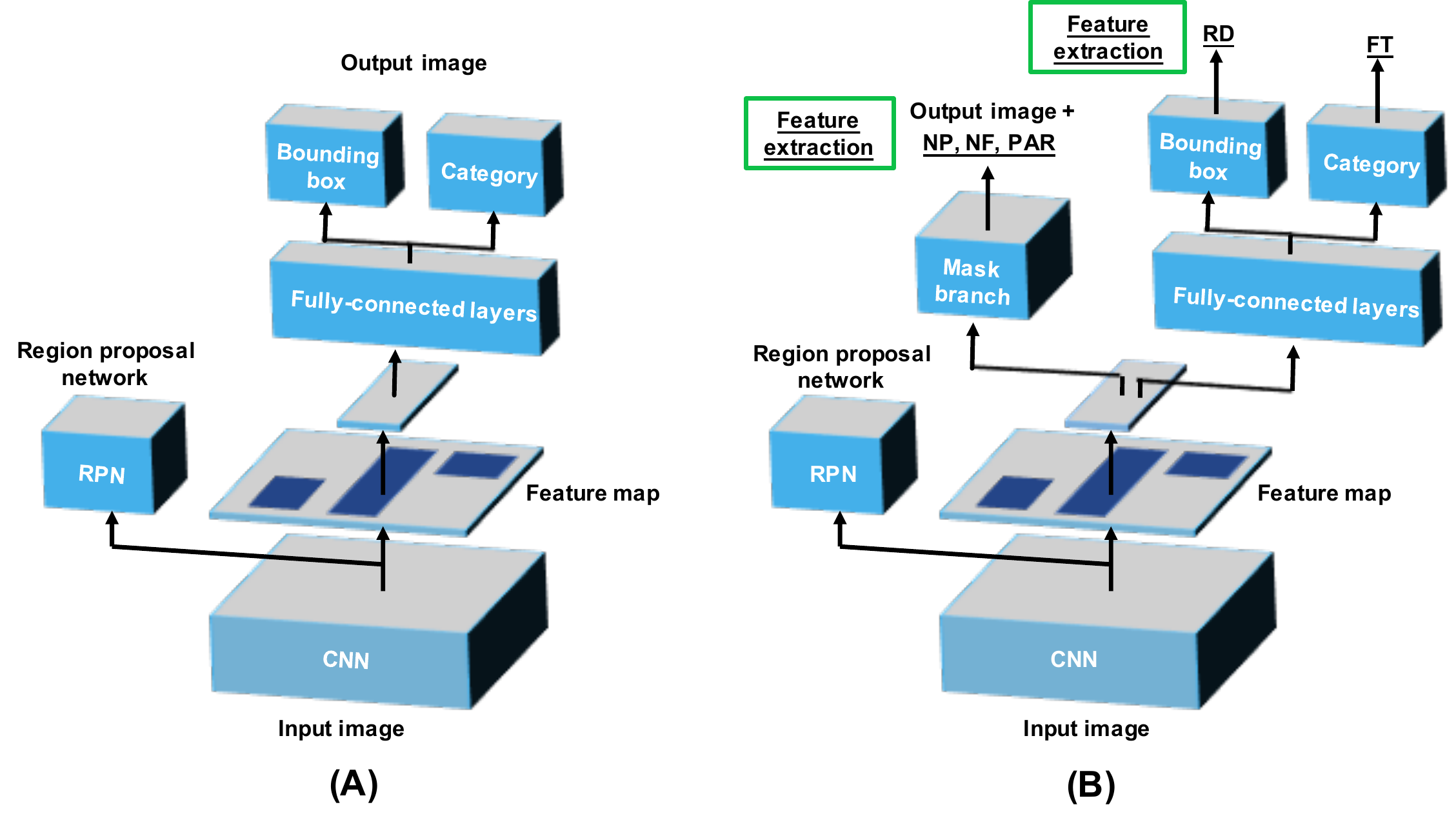}
\caption{(A) Prior work implemented using Faster R-CNN (B) Our proposed pipeline implemented using Mask R-CNN, with handcrafted features extracted from the segmentation masks for consumed food weight estimation}
\label{maskrcnn1}
\end{figure}

\begin{figure}[tb]
\centering
\includegraphics[width=\columnwidth]{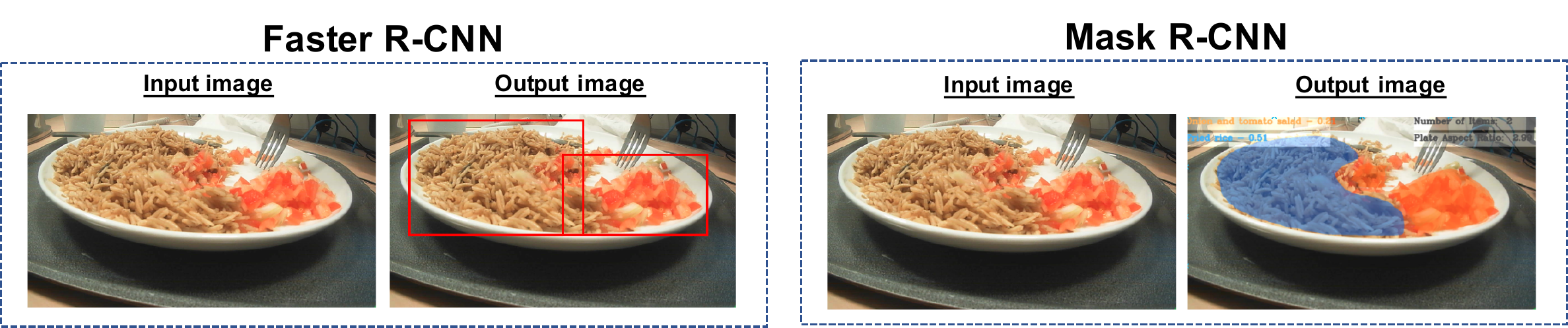}
\caption{Images showing the different food weight estimation methods; bounding boxes applied on input images using Faster R-CNN (left image) and segmentation masks applied to input images using Mask R-CNN (right image)}
\label{maskrcnn2}
\end{figure}

As shown in Figure \ref{detailedpipeline}, the automatic food intake assessment is carried out using a three-stage approach: (a) semantic segmentation, (b) feature extraction and (c) consumed food weight estimation. Instead of using traditional segmentation method, such as local variation \cite{he2013food} and GrabCut \cite{rother2004grabcut}, to extract food items segments, we take the advantage of the promising ability of deep convolution neural network to detect and recognise the food items. A similar idea was reported in a recently published study \cite{jiang2020deepfood}, which applies Faster R-CNN model to generate multiple region proposals, i.e., bounding boxes, on the input food images to achieve dietary assessment. However, the bounding box only indicates the location of the food items and little information about the portion size is disclosed using the method \cite{meyers2015}. As a result, the bounding box method is just based on the assumption of a basic weight, e.g., an assumption that a bounding box can only contain 400g of a food item. In our methods, we further extend the work to generate segmentation masks of food items using Mask R-CNN model, rather than just the bounding boxes obtained through Faster R-CNN model, as shown in Figure \ref{maskrcnn1} and \ref{maskrcnn2}. The loss function of Mask R-CNN can be shown as follows:
\begin{equation}
\mathcal{L}_{total}=\mathcal{L}_{cls}+\mathcal{L}_{box}+\mathcal{L}_{mask}
\end{equation}
where $\mathcal{L}_{cls}$, $\mathcal{L}_{box}$ and $\mathcal{L}_{mask}$ refer to the loss of classification, localization and segmentation mask respectively. $\mathcal{L}_{cls}$ and $\mathcal{L}_{box}$ are same as in Faster R-CNN model. $\mathcal{L}_{mask}$ can be represented as the average binary cross-entropy loss and its formula for food category $k$ can be described as follows \cite{he2017mask}:
\begin{equation}
\mathcal{L}_{mask}=-\frac{1}{d^2}\sum_{1\leq i,j\leq d}[y_{ij}log\hat{y}_{ij}^k+(1-y_{ij})log(1-\hat{y}_{ij}^k)]
\end{equation}
where $y_{ij}$ is the ground-truth label of the pixel(i,j) in a region of size $d\times d$ and $\hat{y}_{ij}^k$ is the predicted label of the corresponding pixel learned for food category $k$. 

When the loss function $\mathcal{L}_{total}$ is minimized, the segmentation masks of food items and containers are generated. By doing so, the segmentation masks help extract handcrafted features for each particular food item and facilitate consumed food weight estimation for the following procedure. Compared to their method, our approach provides more spatial details on the food items. The Mask R-CNN used in our study is implemented by \cite{abdulla2017mask}, which was pre-trained on the COCO dataset \cite{lin2014microsoft}. However, due to the difference between the food categories in the COCO dataset and African food items, the Mask R-CNN can only segment a limited number of African food categories and which is not sufficient to develop a food intake assessment system for use in LMIC. Therefore, 15 common African food items, shown in Table \ref{tab:category}, are chosen and annotated using VGG Image Annotator. It is worth noting that we also label the food remnants so that the segmentation masks can also be obtained after eating. Apart from food categories, containers (e.g., plate, bowl) are annotated to an individual category to facilitate both the data preparation and food weight estimation. Then the resulting images are merged with a subset of Diabetes60 dataset \cite{Christ_2017_ICCV} which contains mainly the \textit{Container} class, as shown in Table \ref{tab:category}, to help locate Region of Interest (ROI) of the images. Since the food items in Diabetes60 dateset are not from African regions which is beyond the scope of this paper, all food items in Diabetes60 dataset are annotated as \textit{Other food} class. In total, 10$k$ images with their corresponding segmentation masks are prepared for training. Data augmentation has also been used to significantly increase the diversity of the dataset for training the models, in which flipping, rotating, shearing and cropping are applied to input images. All the networks are trained using Adam optimiser for 3k epochs with the batch size of 20. The best-trained model is then applied to the images after applying data preparation algorithms for feature extraction. Unless otherwise stated, we train our models with train:validation:test sets of a 10:0.5:1 split. Note that the model evaluation for semantic segmentation will not be presented in this paper since we want to focus more on the food weight estimation task.

\begin{table}[b]
\caption{Selected categories for fine-tuning Mask R-CNN }
\centering
\label{tab:category}
\resizebox{\columnwidth}{!}{%
\begin{tabular}{ccccc}
\hline
\multicolumn{1}{l}{\textbf{Food Dataset}} & \multicolumn{4}{c}{\textbf{Category}} \\ \hline
\multirow{4}{*}{Our dataset \cite{jobarteh2020development}} & Onions \& tomato salad & Tilapia fish & Ugali & Yam \\
 & Chicken drumstick & Spinach stew & Avocado & Banku \\
 & Tomato soup & Roasted beef & Chapati & Onions \\
 & Fried rice & Salted fish & Beef stew & Container \\ \hline
\multicolumn{1}{l}{Diabetes60 \cite{Christ_2017_ICCV}} & Container & Other food &  &  \\ \hline
\end{tabular}%
}
\end{table}

\begin{figure*}[t]
\centering
\includegraphics[width=\textwidth]{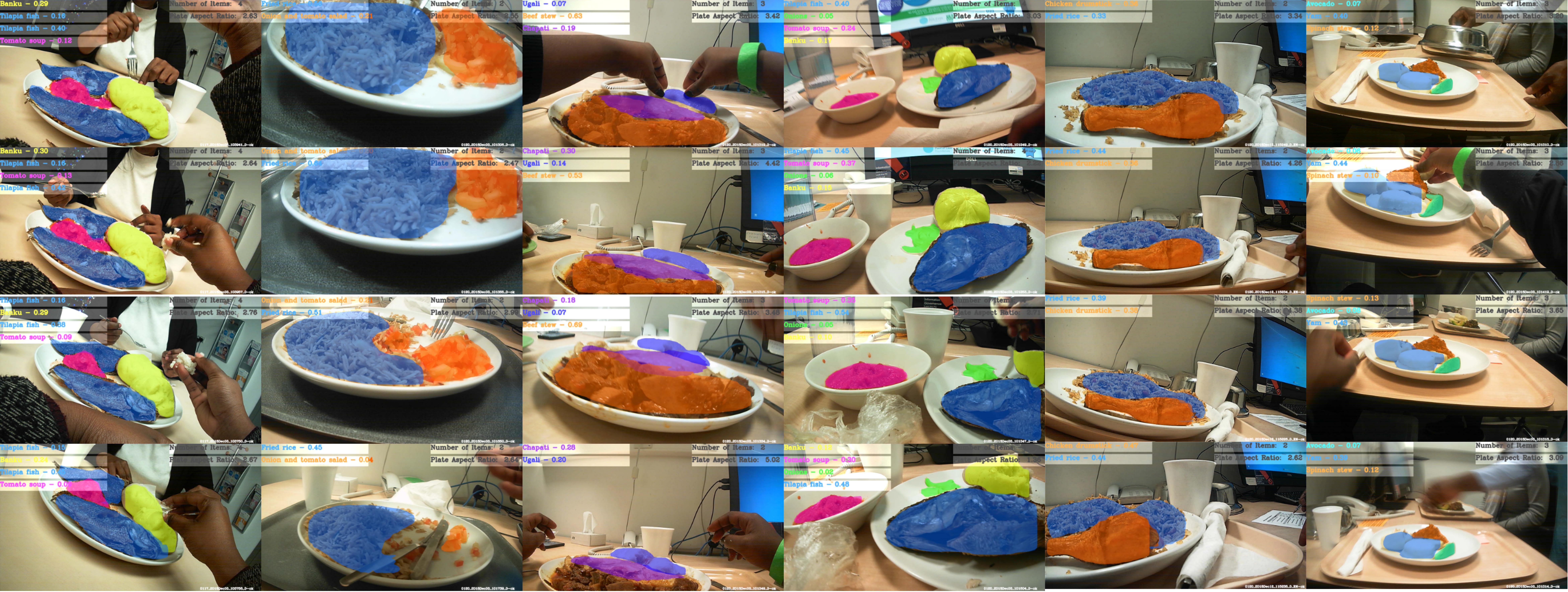}
\caption{Some samples of the experimental results obtained through the semantic segmentation implemented by the Mask R-CNN model, with the details of handcrafted features presented for each frame; FRR (top left), PAR (top right)}
\label{fig:ebutton}
\vspace{-10pt}
\end{figure*}

\subsection{Feature extraction for weight estimation}
Despite a good success in food recognition and segmentation by deep neural networks, there still exists a wide range of hurdles in deep learning-based food portion size estimation \cite{lo2019point2volume}. First, there are insufficient representative training data to train a portion size estimation network since the portion size annotation is complicated, in which either a standardised weighing scale is used to weight each particular food item or water displacement method is applied to measure the food volume. Both methods are labour intensive, as a result there are no existing datasets with annotated portion size sufficient to train an end-to-end deep neural network. Consequently, implicit feature extraction using deep neural network becomes difficult and inefficient. An alternative method is proposed by Google \cite{meyers2015} recently to estimate the food volume in the absence of volume annotation based on deep learning. \cite{meyers2015} makes use of the coupled nature of depth and volume, and replaces volume annotation by depth annotation which can be easily obtained through stereo/depth cameras, e.g., RealSenseF200 depth sensor. Their model is pre-trained on NYU v2 RGBD dataset and then fine-tuned on GFood3d with 150k frames from various Google cafes. Despite the convenience of the strategies, the performance is not yet satisfactory and depth estimation in a non-lab setting fails easily. To overcome these problems, we proposed an efficient and robust strategy which is to train a network based on newly designed handcrafted weight-related features for food weight estimation. It is known that trained dietitians are able to estimate visually the quantity of food on images without relying on any measuring devices. In close cooperation with experienced dietitians, we turn their derived insights on visual food weight estimation into features to help facilitate the solving of the problem \cite{jobarteh2020development}. By using this approach, less representative training data is required since the estimation does not rely on hidden weight-related features anymore. Another advantage is that the network becomes more robust in handling unseen scenarios in the wild which facilitates the food weight estimation in LMIC. The detailed information of the features are summarised as follows:

\begin{itemize}

\item \textbf{Food Type (FT):} One-hot encoding is applied in which each categorical feature is encoded as a one-hot numeric array. It is represented by a vector with all 0 except one, which has 1 as value in its corresponding food category ($15\times1$ vector in this case due to 15 food types).
\item \textbf{Food Region Ratio (FRR):} This feature refers to the ratio of the pixel number of each individual food item to plate ($NF_i/NP$), which gives an indication of the proportion of the region for each food item in the container. 
\item \textbf{Pixel Number of the Plate (NP):} The feature gives a rough estimation of the plate size in the absence of a reference object or fiducial marker.
\item \textbf{Plate Aspect Ratio (PAR):} This represents the long width ratio of the container, which gives an indication of the camera viewing angle. \textit{PAR} can be derived using the coordinates of the plate's segmentation masks based on Singular Value Decomposition (SVD), as shown in the following equation and in Figure \ref{fig:svd}.

\begin{figure}[b]
\centering
\includegraphics[width=1\columnwidth]{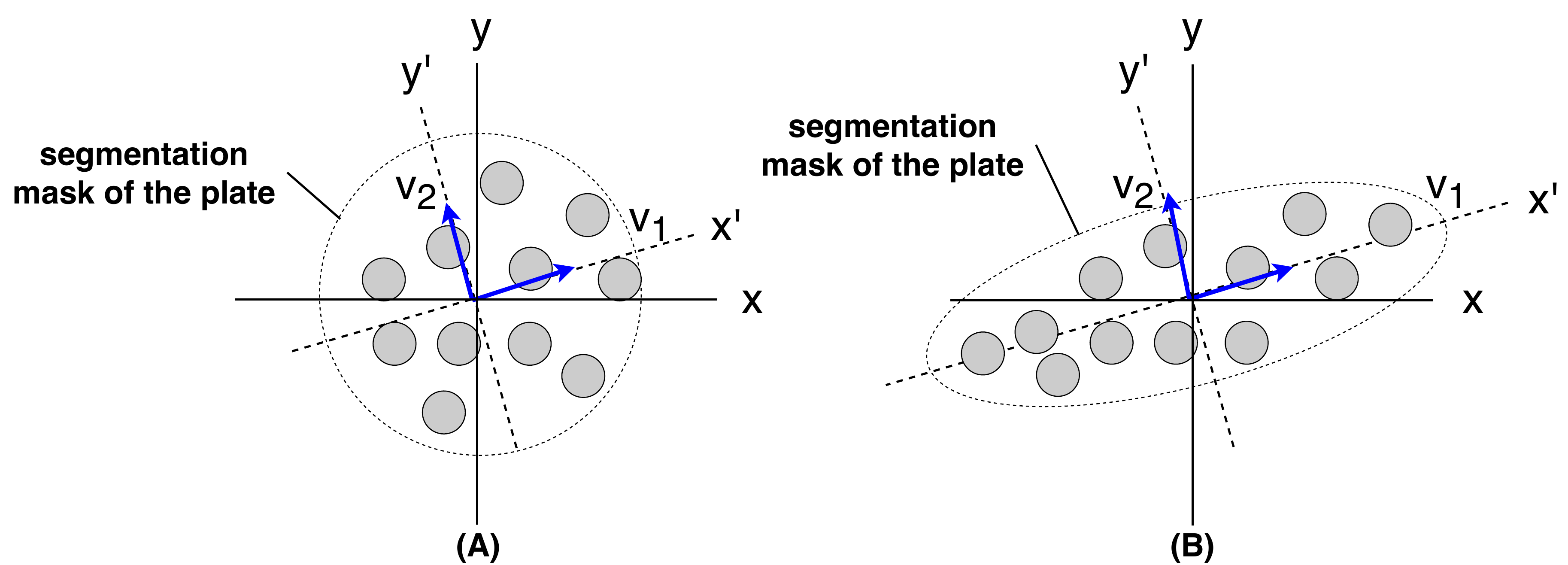}
\caption{Singular Value Decomposition (SVD) on the segmentation mask of the plate (A) the camera is tilted largely downward (e.g., top-down view) (B) the camera is tilted slightly downward (e.g., more horizontal) }
\label{fig:svd}
\end{figure}

\vspace{7pt}

\newcommand{\vect}{\mathbf}
\newcommand{\nul}{\operatorname{Nul}}
\newcommand{\col}{\operatorname{Col}}
\newcommand{\row}{\operatorname{Row}}

\resizebox{\linewidth}{!}{
$X= U\Sigma V^T=
  \begin{matrix}
    \underbrace{\left[\begin{matrix}\vect u_1 & \vect u_2 & \dots & \vect u_r\end{matrix}\right.}& 
    \underbrace{\left.\begin{matrix}\vect u_{r+1} & \dots &  \vect u_m\end{matrix}\right]}\\
    \col X & \nul X^T
  \end{matrix}
  \begin{bmatrix}
      \boxed{\sigma_1} & 0 & \dots & 0 & 0 & \dots & 0 \\
         0 & \boxed{\sigma_2}  & \dots & 0 & 0 & \dots & 0 \\
         \dots& & & & &  \\
         0 & 0 & \dots & \sigma_r  & 0 & \dots & 0 \\
         0 & 0 & \dots & 0 & 0 & \dots & 0 \\
         \dots& & & & &  \\
         0 & 0 & \dots & 0 & 0 & \dots & 0 
  \end{bmatrix}
  \begin{bmatrix}
    \vect v_1^T \\ \vect v_2^T \\ \dots \\ \vect v_r^T \\
    \vect v_{r+1}^T \\ \dots \\ \vect v_n^T
  \end{bmatrix}
  \begin{matrix}
    \left.\vphantom{\begin{bmatrix}
       \vect v_1^T \\ \vect v_2^T \\ \dots \\ \vect v_r^T 
       \end{bmatrix}}\right\}\row X \\ 
    \left.\vphantom{\begin{bmatrix}
      \vect v_{r+1}^T \\ \dots \\ \vect v_n^T 
    \end{bmatrix}}\right\}\nul X
  \end{matrix}$}
  \vspace{7pt}

\noindent
where the columns of $U$ and the rows of $V^T$ are the left singular vectors and right singular vectors respectively. $\Sigma$ is a diagonal matrix which presents the singular values. Then the $PAR$ can be calculated by $\frac{\sigma_{1}}{\sigma_{2}}$.

\item \textbf{Area Weight Ratio (AWR)} Different food items have their corresponding geometry as well as density ($g/cm^3$), and these features will affect the food weight estimation. However, from a single image without relying on any depth information, it is often difficult to infer the full geometry of the food items. Thus, \textit{AWR} feature is designed for rough weight estimation using a single image. Each food item has its corresponding \textit{AWR} value, e.g., Beef Stew: \textit{2.3}; Yam: \textit{3.5}; Avocado: \textit{2.0}, which represents the amount of weight per $cm^2$ ($g/cm^2$). This value can be determined through regression for each particular food item before the experiments (i.e., draw a plot of mean surface area ($cm^2$) versus weight ($g$) of each food item).

\item \textbf{Relative Distance (RD)} During passive monitoring, the position of a food item in an image affects its weight estimation. Food items at the back of an image in a horizontal viewing angle have a larger volume and weight even though the number of pixel is the same as the food items at the front regarding to the pinhole camera model. \textit{RD} roughly indicates the relative position of the food items on a container. It groups food items depending on their position on plate; front and rear facing food items are assigned to different groups (0 or 1).
\end{itemize}

\subsection{Feature importance}

\begin{figure}[htb]
\centering
\includegraphics[trim=100 400 350 550,clip,width=\columnwidth]{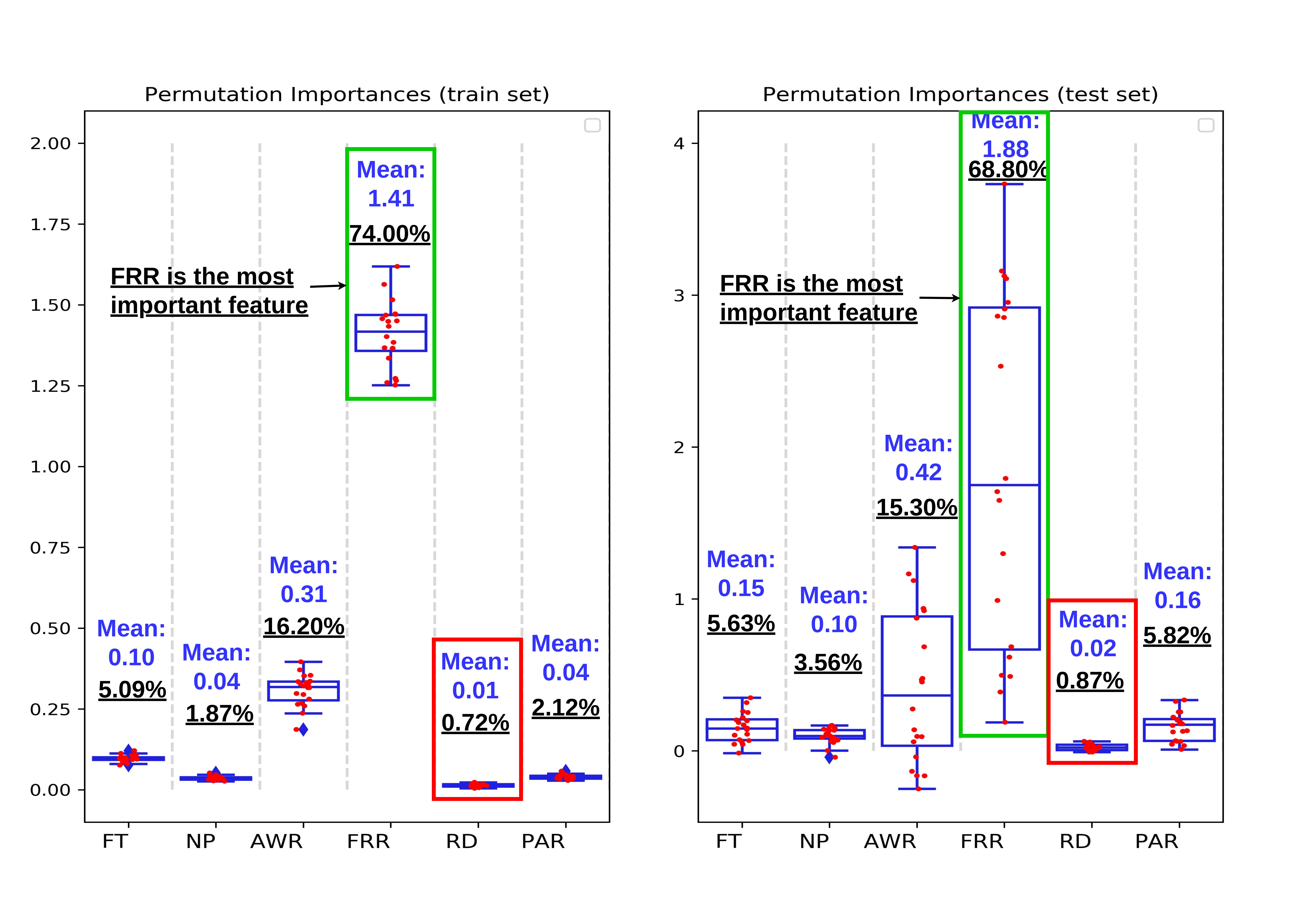}
\caption{Permutation importance of training and test dataset}
\label{fig:permutation}
\end{figure}

In designing an effective weight estimation model, it is equally important to not only estimate the food weight accurately, but also develop an interpretable model, i.e., apart from knowing what the estimated weight is, we wonder why the weight is high/low and which are the most important features in determining the estimation. To the best of our knowledge, we are the first group using handcrafted features to achieve food weight estimation. It is for these reasons that we need to evaluate the feature importance of our proposed features and figure out how these features contribute to the final estimation. Mean Decrease Accuracy (MDA), also known as permutation importance \cite{breiman2001random}, is used to validate the performance of the designed feature. Permutation importance is a model inspection method which is especially useful for non-linear estimators in which the concept is to break the relationship between the target and each particular feature by random shuffling across different subjects. Then the drop in estimation result is indicative of how much the model depends on this particular feature. The equation can be shown as follows:
\begin{equation}
    i_{k} =s-\frac{1}{M}\sum_{m=1}^{M}s_{m,k}
\end{equation}
\noindent
where $i_{k}$ refers to the permutation importance for feature $k$ and $M$ refers to the number of repetition ($M$=20). $s$ represents the reference score of the model evaluated on raw dataset, while $s_{m,k}$ refers to the reference score of the model evaluated on the shuffled dataset (for feature $k$). Figure \ref{fig:permutation} shows that all features have positive permutation importance while a positive value corresponds to a deviation from the null hypothesis (i.e., null hypothesis refers to no statistically significant relationship between the feature and the weight). In both training and test dataset, we can easily observe that the weight estimation model relies heavily on FRR (i.e., $68.8\%$ as shown in Figure \ref{fig:permutation}). This shows the significance of the proportion of food region and indicates the necessity for proposing data preparation algorithms. Apart from FRR, AWR also contributes significantly to the weight estimation (i.e., $15.3\%$ as shown in Figure \ref{fig:permutation}), which is explainable because of the coupled nature of food density and weight. For the others (i.e., FT, NP and PAR), each of them is also able to produce a reasonable average contribution of $5.0\%$. Despite the small contribution of RD, it is worth noting that it also facilitates the weight estimation after comparing with the model trained without RD. Detailed information and experimental results can be found in the following section.

\begin{figure}[tb]
\centering
\includegraphics[width=\columnwidth]{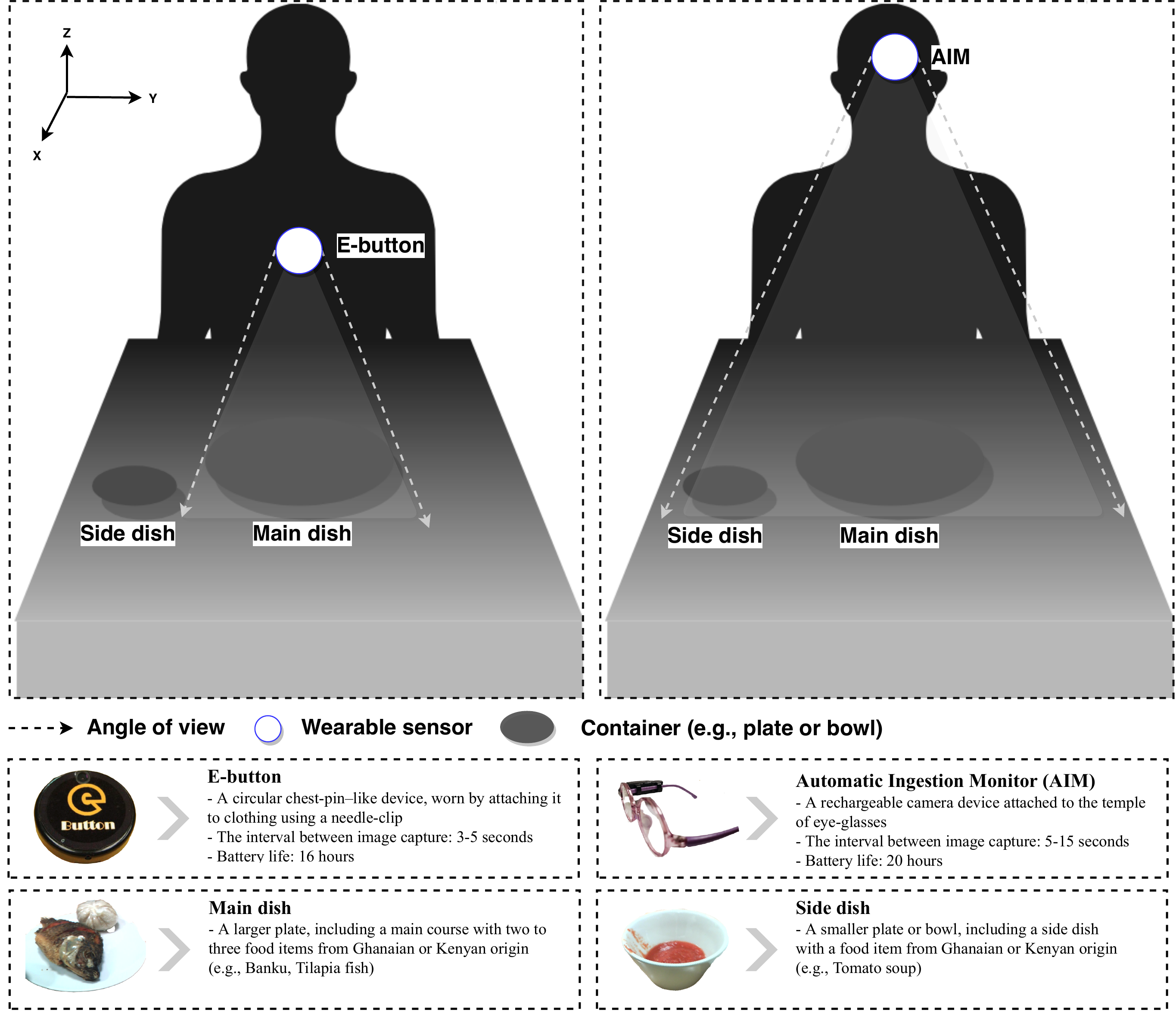}
\caption{The experiment setup and the study protocol of the passive food intake assessment system}
\label{fig:setup}
\end{figure}

\begin{figure}[htb]
\centering
\includegraphics[width=1\columnwidth]{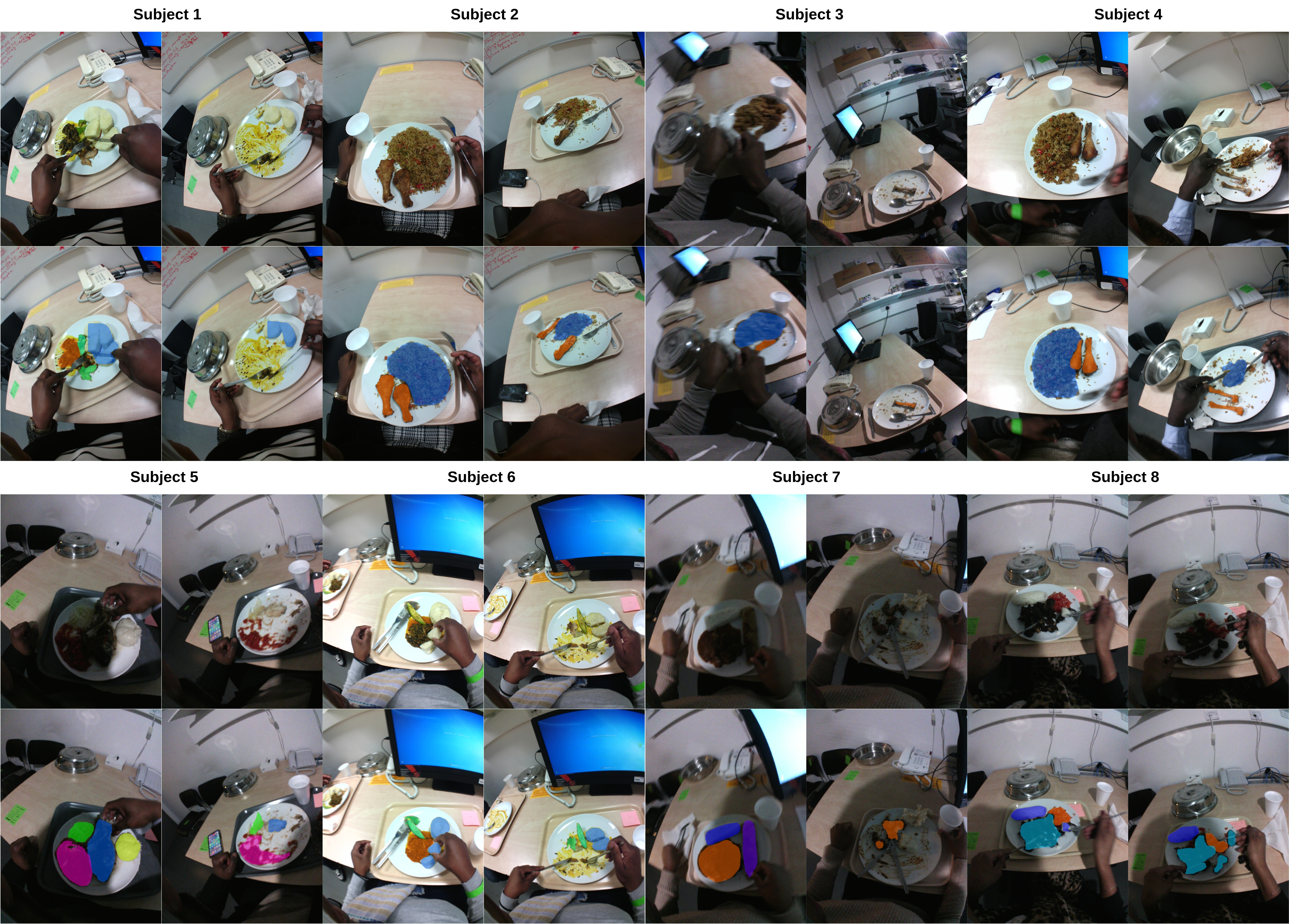}
\caption{Some samples of different eating scenarios with images captured by AIM before and after the meals}
\label{fig:aim}
\end{figure}

% \usepackage{graphicx}
\begin{table*}[t]
\centering
\caption{Quantitative results of food weight estimation using different machine learning approaches}
\label{tab:weightestimation}
\resizebox{\textwidth}{!}{
\begin{tabular}{lccccccc}
\hline
\multicolumn{1}{c}{\textbf{\begin{tabular}[c]{@{}c@{}}Abbreviation of \\ the features\end{tabular}}} & \textbf{\begin{tabular}[c]{@{}c@{}}Number of \\ features\end{tabular}} & \textbf{Classifier} & \textbf{\begin{tabular}[c]{@{}c@{}}Weight estimation \\ error (g) in MAE\end{tabular}} & \textbf{\begin{tabular}[c]{@{}c@{}}Standard \\ Deviation\end{tabular}} & \textbf{\begin{tabular}[c]{@{}c@{}}Weight estimation\\  error (g) in RMSE\end{tabular}} & \textbf{\begin{tabular}[c]{@{}c@{}}Standard \\ Deviation\end{tabular}} & \textbf{\begin{tabular}[c]{@{}c@{}}Accuracy\\ (\%)\end{tabular}} \\ \hline
FRR & 1 & Support Vector Regression (SVR) & 61.03 & 25.75 & 80.81 & 29.33 & 64.12 \\
FRR-FT & 16 & Support Vector Regression (SVR) & 61.96 & 18.23 & 81.48 & 22.10 & 56.03 \\
FRR-FT-NP-AWR-RD-PAR & 20 & Support Vector Regression (SVR) & 57.20 & 15.03 & 76.18 & 21.93 & 65.71 \\ \hline
FRR & 1 & MultiLayer Perceptron (MLP) & 59.50 & 18.27 & 80.09 & 23.39 & 62.85 \\
FRR-FT & 16 & MultiLayer Perceptron (MLP) & 55.30 & 33.74 & 72.25 & 36.61 & 62.69 \\
FRR-FT-NP-AWR-RD-PAR & 20 & MultiLayer Perceptron (MLP) & 60.54 & 26.90 & 84.67 & 41.23 & 64.02 \\ \hline
FRR & 1 & Random Forest (RF) & 62.76 & 26.03 & 78.50 & 27.20 & 52.53 \\
FRR-FT & 16 & Random Forest (RF) & 46.36 & 20.62 & 63.01 & 22.38 & 69.36 \\
FRR-FT-NP-AWR-RD-PAR & 20 & Random Forest (RF) & \underline{\textbf{40.19}} & 16.58 & \underline{\textbf{51.73}} & 20.58 & 72.38 \\ \hline
FRR & 1 & Extra Trees (ET) & 64.82 & 29.21 & 82.23 & 32.18 & 53.49 \\
FRR-FT & 16 & Extra Trees (ET) & 46.78 & 21.42 & 61.80 & 21.89 & 65.71 \\
FRR-FT-NP-AWR-RD-PAR & 20 & Extra Trees (ET) & 43.07 & 19.33 & 54.30 & 21.30 & 70.95 \\ \hline
FRR & 1 & Gradient Boosted (GB) & 61.73 & 23.26 & 76.76 & 28.26 & 57.46 \\
FRR-FT & 16 & Gradient Boosted (GB) & 51.95 & 19.43 & 65.81 & 21.16 & 69.04 \\
FRR-FT-NP-AWR-RD-PAR & 20 & Gradient Boosted (GB) & 48.75 & 19.02 & 61.07 & 22.25 & 72.06 \\ \hline
FRR & 1 & Decision Trees (DT) & 57.57 & 23.78 & 72.19 & 28.13 & 62.85 \\
FRR-FT & 16 & Decision Trees (DT) & 57.57 & 23.78 & 72.19 & 28.13 & 62.85 \\
FRR-FT-NP-AWR-RD-PAR & 20 & Decision Trees (DT) & 53.96 & 21.72 & 67.28 & 25.69 & 66.19 \\ \hline
FRR & 1 & RF+ET+MLP$_{Ensemble}$ & 64.23 & 28.06 & 81.23 & 29.60 & 57.62 \\
FRR-FT & 16 & RF+ET+MLP$_{Ensemble}$ & 49.59 & 22.21 & 65.20 & 24.42 & 65.07 \\
FRR-FT-NP-AWR-RD-PAR & 20 & RF+ET+MLP$_{Ensemble}$ & 42.27 & 15.82 & 54.97 & 22.43 & \underline{\textbf{74.60}} \\ \hline
- & - & Baseline & 92.47 & 21.50 & 110.17 & 23.69 & 34.76 \\ \hline
\end{tabular}

}
\vspace{-10pt}
\end{table*}

\section{Experimental Results and Discussions}

\subsection{Study Design and Data Collection Protocol}

In cooperation with the University of Pittsburgh and University of Alabama, this study aims to evaluate the functionality and acceptability of the custom egocentric cameras, with eButton \cite{sun2014ebutton} and AIM \cite{doulah2020automatic}, and the performance of the proposed pipeline in food intake assessment using the images captured under constrained settings, mimicking special LMIC environments and conditions. These conditions include using food items of Ghanaian and Kenyan origin, eating under low lighting condition to simulate an environment of inadequate supply of electricity, as shown in Figure \ref{fig:setup}. 13
healthy subjects ($\geq$18 years) of Ghanaian or Kenyan origin living in London, United Kingdom, were recruited to participate in the study held at the National Institute for Health Research Clinical Research Facility (CRF) at the Hammersmith Campus of Imperial College London for 3 times\footnote{The Imperial College Research Ethics Committee has provided ethics approval for these studies: approval reference number is 18IC4780 for the study at the CRF; This project is registered at clinicaltrials.gov (NCT03723460)}. Both eButton and AIM will be assigned to different subjects. Before the start, a standardised weighing scale (i.e., Salter Brecknell) is used to first pre-weigh the food items. Preweighed food items are placed on the food containers and presented to the subjects, and the egocentric cameras will be used during eating. Instead of finishing all the food items, the subjects are asked to just eat until full. Then the food remnants are post-weighed and recorded, so that the difference between pre-weighed and post-weighed food items can be considered as the ground truth of the consumed food weight (i.e., also known as weighed food records). In order to demonstrate the efficacy of the proposed system, experienced dietitians are also involved to perform the visual food weight estimation, and their predictions are compared to our proposed method. To ensure the fairness, the accessors are trained to be well-known with the foods of Ghanaian and Kenyan origin and also familiar with the technique of visual food weight estimation.

\begin{figure*}[htb]
\centering
\includegraphics[width=\textwidth]{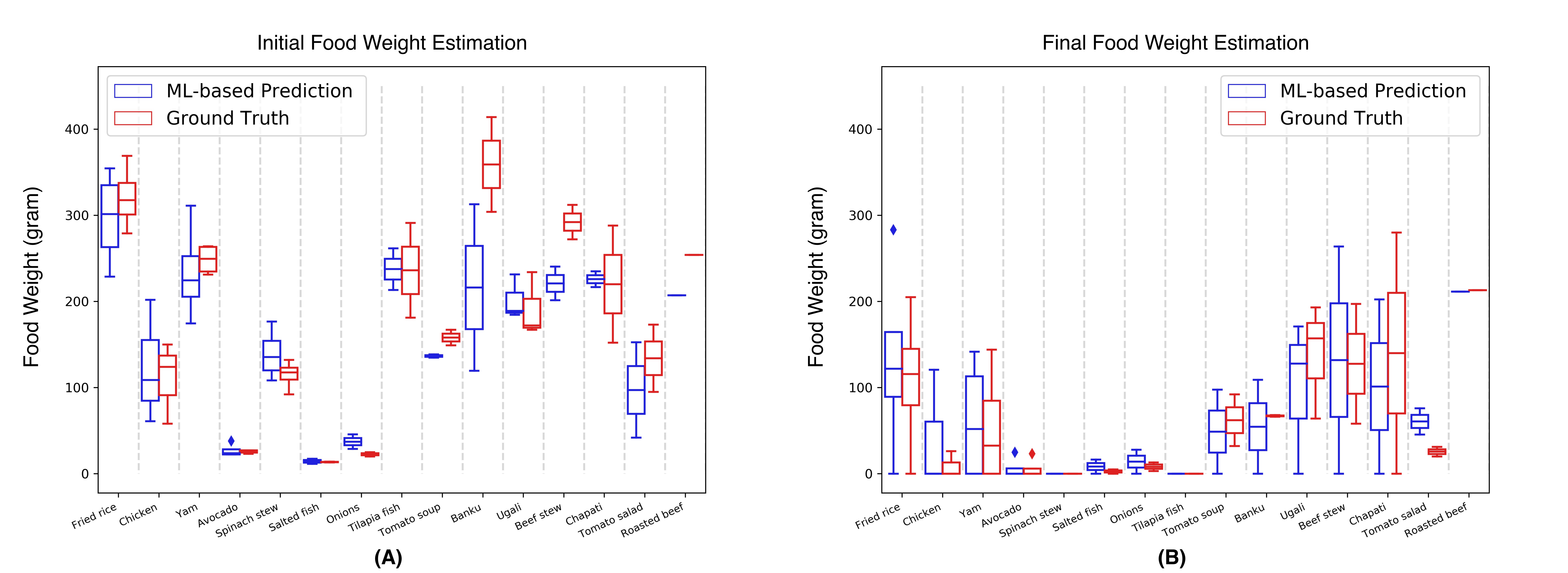}
\caption{Initial and final food weight estimation using ML-based approach}
\label{fig:initialandfinal}
\vspace{-10pt}

\end{figure*}

\subsection{Performance of Mask R-CNN in segmenting food items and extracting features}

The qualitative results for both eButton and AIM are shown in Figure \ref{fig:ebutton} and Figure \ref{fig:aim} respectively. It is encouraging to find that Mask R-CNN has promising results in generating food segments, even for food remnants, as shown in Figure \ref{fig:aim}. In order to determine whether our proposed method is view-invariant, we visualise some images with varying viewing angles in the same scenario in Figure \ref{fig:ebutton} (i.e., most of them are captured before eating except for the $2nd$ column). The value of the extracted handcrafted features, especially for \textit{FRR}, is stable across different frames which can be easily observed in the $1st$ and $6th$ column. From this fact, it is easily concluded that the proposed weight estimation method is robust and efficient in handling images from different positions. It may be also worth mentioning that the \textit{FRR} for the last image in both $3rd$ and $5th$ column is larger compared to the rest of the images in the same scenario, which is because the food containers were poorly captured in those images. The incomplete containers capture increases the proportion of food items in the containers which in turn overestimate the food weight. It is for this reason that those images will be removed using data preparation algorithm before inputting to the estimation network. Another interesting finding is that when we compare the value of \textit{PAR} in the $1st$ column with those in $3rd$ column, we found that the former one is smaller. This is because the images in the $1st$ column are captured by a camera tilted largely downwards (i.e., similar to a top-down view), of which the illustration diagram is already demonstrated in Figure \ref{fig:svd}. To further evaluate the performance of \textit{PAR}, we also train a MLP classifier for a new problem which is to classify whether the image is captured by eButton or AIM using \textit{PAR} as the only feature (i.e., the images captured by AIM tilted largely downwards, while the images captured by eButton are not, as shown in Figure \ref{fig:setup}). The final testing accuracy is $98.5\%$ which shows a promising preliminary result on the challenging problem of estimating viewing angles of the cameras from images captured in the wild without relying on any deep learning-based approaches. These findings show that the viewing angles can be easily inferred through the feature \textit{PAR} (i.e., the aspect ratio of the food containers), which in turn provide valuable insights into how to handle images captured with unseen positions. Note that the assumption is that the food containers are usually round.

\subsection{Performance of weight estimation networks using extracted features}

In Table \ref{tab:weightestimation}, we report weight estimation error in MAE, RMSE and their standard derivation of the models using different machine learning-based approaches trained by the proposed handcrafted features. Apart from these standard metrics, we further calculate the accuracy of the weight estimation, which can be defined as:
\begin{equation}
\frac{1}{M}\sum_{m=1}^{M}\mathcal{F}(\left \| \hat{w}-g \right \|<\epsilon )
\end{equation}

\noindent
where $\mathcal{F}(\texttt{true}) = 1$ and $\mathcal{F}(\texttt{false}) = 0$ is a Boolean function, and $\epsilon$ is the tolerance for determining a correct estimation ($\epsilon=50$). $M$ refers to the total number of test data, $\hat{w}$ and $g$ represents the estimated weight and the ground truth respectively. Unlike previous works which use food volume error as the evaluation metric \cite{lo2019point2volume}, food weight error is used in our work. Following prior works, k-fold cross validation ($k=15$) is used to examine the trained weight estimation models to ensure the fairness and test their robustness. For comparison, we also present a baseline which is simply to estimate the food weight using the median value of the targeted food items for every observation in the test data. Regarding to the feature importance reported in Figure \ref{fig:permutation}, we first show the quantitative result of the model trained using \textit{FRR} as the only feature. It is encouraging to show a promising preliminary result on the estimation which significantly outperforms the baseline model. Afterwards, we add \textit{FT} as a new feature to train another set of weight estimation models. It is worth noting that the \textit{FRR-FT} models achieve superior performance against the \textit{FRR} only models, indicating that food weight estimation relies heavily on food categories. From this finding, we realise the importance of semantic segmentation. Instead of using traditional segmentation method (e.g., GrabCut) in our pipeline to generate food segments without any attempt at understanding what these parts represent, it is reasonable to apply semantic segmentation (i.e., Mask R-CNN) along with ML-based approaches in solving the problem of food weight estimation. Besides, it can be observed that our models with full features achieve the best performance across most of the metrics in different ML-based regressors. Regarding to the great performance of RF and ET in the metrics MAE and RMSE, we further developed an ensemble of RF, ET and MLP trained with all features. We notice that although there is no significant difference in MAE and RMSE with RF and ET, ensemble learning achieves the best accuracy in food weight estimation (i.e., $74.60\%$ as shown in Table \ref{tab:weightestimation}).

\subsection{Performance of weight estimation networks across different food categories}

\begin{figure*}[t]
\centering
\includegraphics[width=\textwidth]{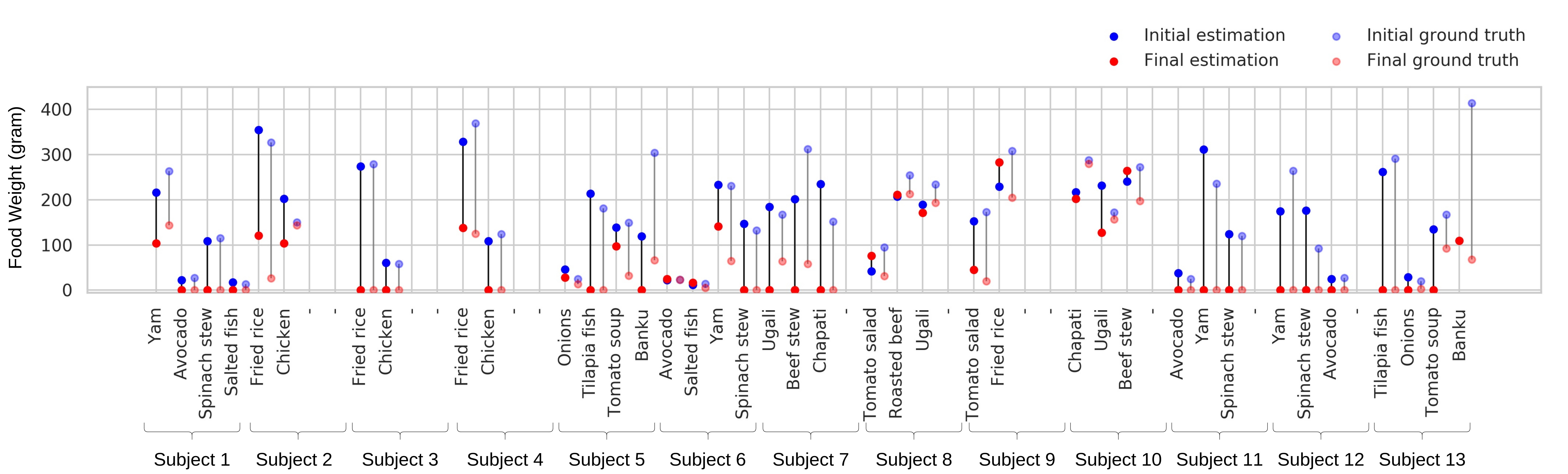}
\caption{Initial and final food weight estimation using ML-based approach}
\label{fig:plotfordifferentsubject}
\end{figure*}

Figure \ref{fig:initialandfinal} shows the experimental results of food weight estimation for each particular food category across different scenarios. Note that the ML-based prediction in this experiment is generated by the ensemble model mentioned above. In Figure \ref{fig:initialandfinal}, it is encouraging to observe that the models can successfully estimate the range of food weight in the scenario before and after the meal even though all images are trained together without separating into two models. It may also be worth mentioning that banku has a larger error in initial food weight estimation (corresponding to pre-weighed food record). This is due to some participants tending to put the uneaten banku on the food tray instead of in the food containers. Thus, our proposed method may inevitably induce weight estimation error since it relies heavily on the proportion between the container and the food items. To better visualise the performance of our proposed system, Figure \ref{fig:plotfordifferentsubject} shows the initial and final food weight estimation across different subjects. The corresponding eating scenario of the subjects can also be referred to the images shown in Figure \ref{fig:aim}. 

\subsection{Comparison of consumed food weight between ML-based and visual weight estimation}

\begin{figure*}[t]
\centering
\includegraphics[width=\textwidth]{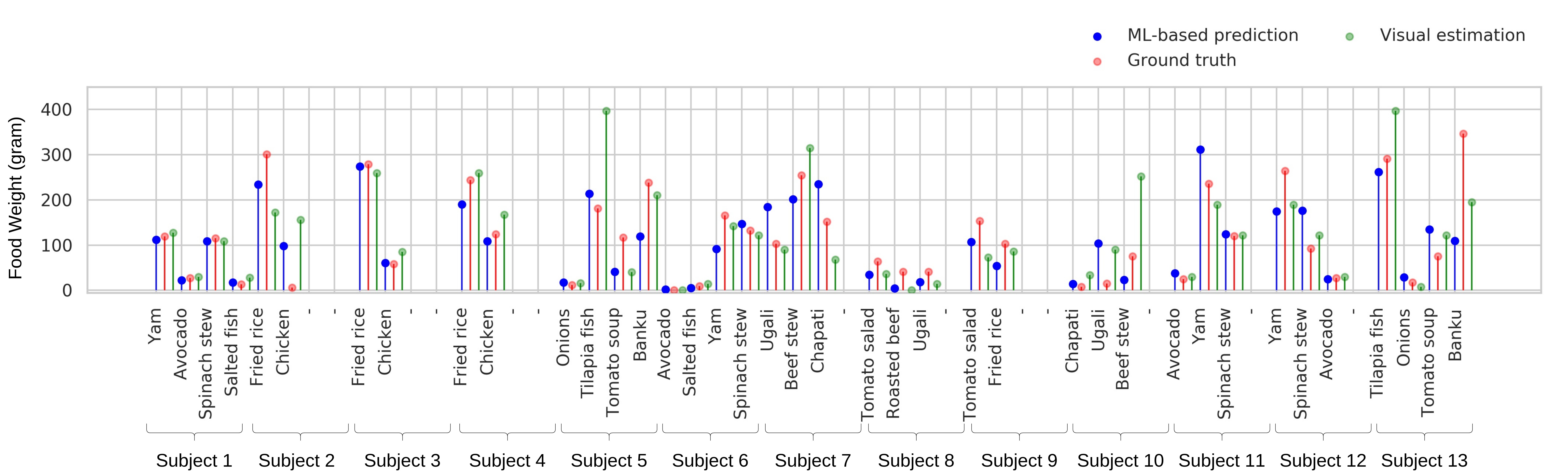}
\caption{Comparison between ML-based approach and visual estimation by experienced dietitians on consumed food weight}
\label{fig:comparewithdietians}
\end{figure*}

Unlike food recognition, the existing research studies on portion size estimation (i.e., either food volume, food calorie or food weight estimation) have only examined their approaches on self-collected in-the-wild test datasets, in which there does not have an existing benchmark to conduct a fair comparison with previous methods. Thus, it is preferable to evaluate our proposed method in terms of practicality and implementation. To evaluate the practicality, we calculate the net difference of the initial and final food weight (i.e., the consumed food weight), and compare with the results estimated visually by the experienced dietitians as shown in Figure \ref{fig:comparewithdietians}. To better visualise the performance, we also report the estimation error of the consumed weight estimation for the AIM, eButton and manual estimation in Table \ref{table:comparison}. We can observe that the overall ML-based approach ($37.6\%\pm27.2\%$) outperforms the manual estimation ($48.8\%\pm63.1\%$). It is also worth noting that the performance of AIM ($32.7\%\pm24.7\%$) is better than eButton ($47.4\%\pm30.4\%$), and this is because AIM is worn at a higher position which can have a better view of the food containers. We also notice that the mean and the standard deviation is relatively large for manual estimation using the image captured by eButton. The reason is probably that the spatial awareness of humans drops significantly when the seen images are captured at a horizontal level (i.e., some food items are occluded).

\begin{table}[t]
\centering
\caption{Estimation error in MAE (\%) of the consumed food weight for the AIM, eButton and manual estimation}
\label{table:comparison}
\resizebox{=0.9\columnwidth}{!}{%
\begin{tabular}{ccccccc}

\hline
 \textbf{Method} & \textbf{AIM} & \textbf{Manual} & \textbf{eButton} & \textbf{Manual} & \textbf{Both} & \textbf{Manual} \\ \hline
Mean (\%) & 32.7 & 36.9 & 47.4 & 72.5 & 37.6 & 48.8 \\
S.D (\%) & 24.7 & 34.0 & 30.4 & 96.3 & 27.2 & 63.1 \\ \hline
\end{tabular}%
}
\end{table}

\section{Future Works}

The proposed food intake assessment system was designed to be used in households in Africa. To ease the procedure in conducting large-scale studies and minimising the effect on the participants' normal behaviour, dietitians require that no reference objects should be used for estimating portion size. In this work, the food containers are usually in standard sizes with small variations. For the scenarios where the food items are placed in unseen containers, this paper has not yet covered. It is for this reason that a more advanced system will be developed for the next stage which gives a rough estimation on the dimension of the containers before semantic segmentation and feature extraction \cite{jia2020estimating}. From the perspective view of hardware design, it is known that time-of-flight camera is efficient in estimating the scale and the size of the containers without relying on any reference objects \cite{lo2019point2volume}. An integrated system based on a wearable depth camera along with the proposed pipeline in food intake assessment could be one of the future directions. In addition, we also plan to combine our vision-based system with existing passive wearable technology, i.e., EMG sensors \cite{zhang2017monitoring} and accelerometers \cite{doulah2020automatic, farooq2018accelerometer}, to further improve the performance of food weight estimation. Last but not least, due to
continuous capture, privacy is also an important concern for passive food intake assessment systems \cite{doulah2020automatic}. Apart from removing non-food images, more efforts will be made, e.g., removal of background people using deep learning-based approaches, to improve the acceptability and feasibility of the system.

\section{Conclusion}

Understanding the quantity and nutritional content of habitual diets eaten at the community, household and individual level is important to reduce malnutrition in LMIC. Thus, an intelligent passive food intake assessment system based on data-driven semantic segmentation and food weight estimation network is presented. By using the proposed system, we can estimate the consumed food weight continuously instead of just measuring the portion size of the whole meal. After comparing with the manual estimation carried out by experienced dietitians, we found that our proposed method not only outperforms traditional measure but also eases the procedure in conducting large-scale dietary assessments. The experimental results also prove that our proposed method can reliably monitor food intake and give feedback on users' eating behaviour which provides guidance for dietitians.

% or

\section*{References}
\renewcommand*{\bibfont}{\footnotesize}
\printbibliography[heading=none]

\end{document}